\documentclass[sigconf]{acmart}

\usepackage{booktabs} % For formal tables
\usepackage{xcolor}

\def\BibTeX{{\rm B\kern-.05em{\sc i\kern-.025em b}\kern-.08em
    T\kern-.1667em\lower.7ex\hbox{E}\kern-.125emX}}

\copyrightyear{2021}
\acmYear{2021}
\setcopyright{rightsretained}
\acmConference[GECCO '21 Companion]{2021 Genetic and Evolutionary Computation Conference Companion}{July 10--14, 2021}{Lille, France}
\acmBooktitle{2021 Genetic and Evolutionary Computation Conference Companion (GECCO '21 Companion), July 10--14, 2021, Lille, France}\acmDOI{10.1145/3449726.3459585}
\acmISBN{978-1-4503-8351-6/21/07}

\begin{document}
\title[A complementarity analysis...]{A Complementarity Analysis of the COCO Benchmark Problems and Artificially Generated Problems}

\author{Urban Škvorc}
\affiliation{Computer Systems Department \\
  Jožef Stefan Institute\\
Jožef Stefan International \\
Postgraduate School \\
Ljubljana, Slovenia \\
urban.skvorc@ijs.si}

\author{Tome Eftimov}
\affiliation{Computer Systems Department \\
Jožef Stefan Institute\\
Ljubljana, Slovenia,\\
tome.eftimov@ijs.si}

\author{Peter Korošec}
\affiliation{Computer Systems Department \\
Jožef Stefan Institute\\
Ljubljana, Slovenia,\\
peter.korosec@ijs.si}

\begin{abstract}

When designing a benchmark problem set, it is important to create a set of benchmark problems that are a good generalization of the set of all possible problems. One possible way of easing this difficult task is by using artificially generated problems. In this paper, one such single-objective continuous problem generation approach is analyzed and compared with the COCO benchmark problem set, a well know problem set for benchmarking numerical optimization algorithms. Using Exploratory Landscape Analysis and Singular Value Decomposition, we show that such representations allow us to further explore the relations between the problems by applying visualization and correlation analysis techniques, with the goal of decreasing the bias in benchmark problem assessment.

\end{abstract}




\maketitle

\section{Introduction}

The first step to creating a good benchmarking environment is to select a good set of benchmark problems. Ideally, a well designed problem set would be representative of the entire set of problems that the benchmark aims to estimate the performance of. For a benchmark set that aims to determine a general performance of a given algorithm, this would mean that the benchmark problems should be evenly distributed over the space of all possible problems that this algorithm will be solving. However, it can be hard to determine what the actual possible problem space even is, and if benchmark problems are created by hand, it can be difficult to create a large enough number of varied benchmark problems to cover the entire problem space.

In this paper, we attempt to ease this task by using an artificial problem generator developed by Tian et al. in \cite{tian_recommender_2020}, and compare these problems to the 24 noiseless benchmark problems of the well known COCO \cite{hansen2020coco} benchmarking platform (COCO problems). The goal of this paper is to gain further knowledge on the COCO problems by analyzing whether and how the artificially generated problems complement the COCO problems. Or if the alternative is true, and these two sets of problems instead form a single group of problems.  We perform this analysis using Exploratory Landscape Analysis (ELA), a methodology that allows us to describe problems using numerical descriptors called landscape features.

In this paper, we built upon our prior work described in \cite{SKVORC2020106138} and \cite{eftimov2020linear}. In addition, we employ a benchmark problem generator described in \cite{tian_recommender_2020}. The work presented in this poster is similar to several existing papers, primarily \cite{munoz2020generating, limit}, but uses a different problem generation approach, as well as a different methodology for visualizing problems.

\section{Methodology}
\label{section:3:methodology}

Our methodology can broadly be split into four steps: the problem selection (including generation), ELA feature calculation, Singular Value Decomposition (SVD) mapping, and finally the complementarity analysis of the resulting SVD representations.

In the first step, we select the problems used for our analysis. The two problem sets used are the 24 noiseless COCO problems \cite{hansen:inria-00362633}, and the set of 500 artificially generated problems using the method described in \cite{tian_recommender_2020}. The problems are generated using the dimensionality $D=10$, and are calculated using a sample size of $200D$.

In the second step, we use ELA to calculate the landscape features that will be used to compare the problems from the two problem sets. The landscape features used are the same as in \cite{SKVORC2020106138}.

In the third step, these landscape features are scaled to values between 0 and 1 using min-max scaling and then transformed into a subspace using SVD as described in \cite{eftimov2020linear} in order to improve the reliability of our results. The values for the COCO and the generated problems are transformed separately. The values of one set are then projected to the SVD subspace of the other. We use three different projection methods: projecting the COCO problems into the SVD subspace of the generated problems, projecting the generated problems into the SVD subspace of the COCO problems, and calculating the SVD representation of both the COCO and generated problems together without projections.

In the fourth step, we use the SVD problem representations obtained in the third step to analyze the complementarity of the two problem sets. To accomplish this, we visualize the benchmark problems in a 2D space using t-SNE as described in \cite{SKVORC2020106138} and perform Pearson correlation analysis.

\section{Results}
\label{section:5:results}
\begin{figure}[htbp]
    \centering
    \includegraphics[width=0.4\textwidth]{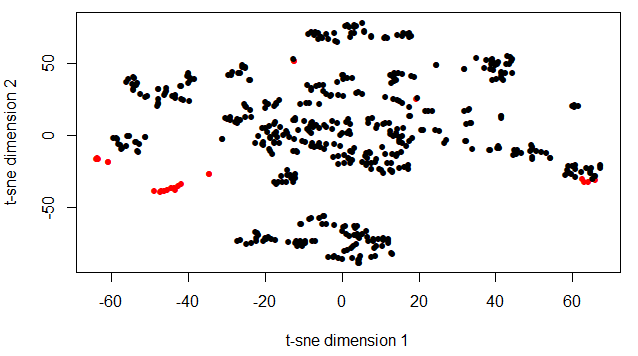}
    \caption{A t-SNE visualization of projecting the set of generated problems (black) into the SVD subspace of the set of COCO problems (red). The COCO problems are distinct from the generated problems.}
    \label{figures:results:vis:coco_to_random}
\end{figure}

Figure \ref{figures:results:vis:coco_to_random} shows the results of projecting the 24 COCO problems into the space of the 500 generated problems. We can see that the COCO problems do not cover the entire space of the generated problems. In particular, there is a large cluster of COCO problems at the left of the visualization that is distinct from the generated problems, as well as a smaller cluster at the right.  The other two projection methods types showed similar results.

\begin{figure}[htbp]
    \centering
    \includegraphics[width=0.4\textwidth]{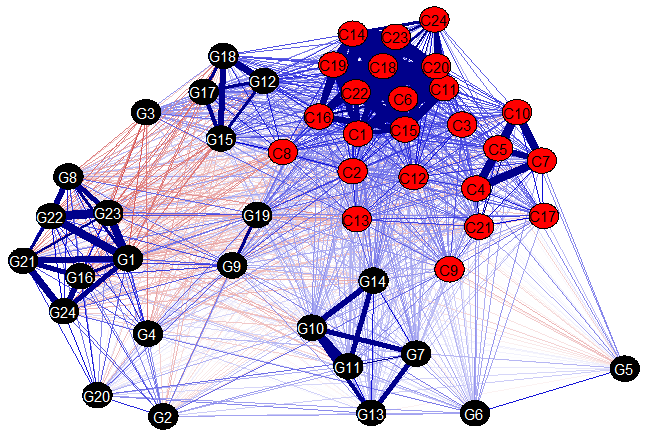}
    \caption{Pearson correlation between the generated problems (black) and the COCO problems (red), after the generated problems have been projected to the SVD subspace of the COCO problems.}
    \label{figures:results:corr:random_to_coco_cos}
\end{figure}

Figure \ref{figures:results:corr:random_to_coco_cos} shows the results of the Pearson correlation analysis. The thickness of the edges shows how correlated a pair of problems are, and its color shows whether the correlation is positive (blue) or negative (red). We can see that the two sets of problems are visually distinct from one another.

\section{Discussion \& Conclusions}
\label{section:6:discussion}
In this paper, we presented the results of a complementary analysis between the commonly used COCO benchmark set and the problems generated by an artificial problem generator.

The foremost conclusion drawn from this paper is that, as far as ELA landscape features are concerned, the 24 COCO benchmark problems represent only a small subset of all possible optimization problems. We believe this is an important realization both for the field of benchmarking (in order to reduce bias analysis in performance assessment), as well as for the field of Exploratory Landscape Analysis. For the field of benchmarking, this paper gives an idea of how problem generation combined with ELA can be used to augment existing benchmark problems. Regarding ELA, we believe these results show that only using traditional benchmark sets such as the 24 COCO benchmark problems might not be enough for a thorough evaluation of these landscape features, and that additional problems should be used.

This paper also presents opportunities for future work. One way to extend this work would be by including additional benchmark sets, for example the CEC benchmark problem set. Another possible way of extending our work is the inclusion of algorithm performance metrics by examining how ELA features correlate with algorithm performance.

\begin{acks}
  This work was supported by projects from the Slovenian Research Agency: research core funding No. P2-0098, project No. Z2-1867 and young researcher funding No. Pr-08987.

\end{acks}

\bibliographystyle{ACM-Reference-Format}
\bibliography{sample-sigconf} 


\begin{thebibliography}{7}

%%% the Web address.

\ifx \showCODEN    \undefined \def \showCODEN     #1{\unskip}     \fi
\ifx \showDOI      \undefined \def \showDOI       #1{#1}\fi
\ifx \showISBNx    \undefined \def \showISBNx     #1{\unskip}     \fi
\ifx \showISBNxiii \undefined \def \showISBNxiii  #1{\unskip}     \fi
\ifx \showISSN     \undefined \def \showISSN      #1{\unskip}     \fi
\ifx \showLCCN     \undefined \def \showLCCN      #1{\unskip}     \fi
\ifx \shownote     \undefined \def \shownote      #1{#1}          \fi
\ifx \showarticletitle \undefined \def \showarticletitle #1{#1}   \fi
\ifx \showURL      \undefined \def \showURL       {\relax}        \fi
% invisible to TeX
\providecommand\bibfield[2]{#2}
\providecommand\bibinfo[2]{#2}
\providecommand\natexlab[1]{#1}
\providecommand\showeprint[2][]{arXiv:#2}

\bibitem[\protect\citeauthoryear{{Eftimov}, {Popovski}, {Renau}, {Korošec},
  and {Doerr}}{{Eftimov} et~al\mbox{.}}{2020}]%
        {eftimov2020linear}
\bibfield{author}{\bibinfo{person}{T. {Eftimov}}, \bibinfo{person}{G.
  {Popovski}}, \bibinfo{person}{Q. {Renau}}, \bibinfo{person}{P. {Korošec}},
  {and} \bibinfo{person}{C. {Doerr}}.} \bibinfo{year}{2020}\natexlab{}.
\newblock \showarticletitle{Linear Matrix Factorization Embeddings for
  Single-objective Optimization Landscapes}. In \bibinfo{booktitle}{{\em 2020
  IEEE Symposium Series on Computational Intelligence (SSCI)}}.
  \bibinfo{pages}{775--782}.
\newblock
\showDOI{%
\url{https://doi.org/10.1109/SSCI47803.2020.9308180}}


\bibitem[\protect\citeauthoryear{Hansen, Auger, Ros, Mersmann, Tušar, and
  Brockhoff}{Hansen et~al\mbox{.}}{2021}]%
        {hansen2020coco}
\bibfield{author}{\bibinfo{person}{Nikolaus Hansen}, \bibinfo{person}{Anne
  Auger}, \bibinfo{person}{Raymond Ros}, \bibinfo{person}{Olaf Mersmann},
  \bibinfo{person}{Tea Tušar}, {and} \bibinfo{person}{Dimo Brockhoff}.}
  \bibinfo{year}{2021}\natexlab{}.
\newblock \showarticletitle{COCO: a platform for comparing continuous
  optimizers in a black-box setting}.
\newblock \bibinfo{journal}{{\em Optimization Methods and Software\/}}
  \bibinfo{volume}{36}, \bibinfo{number}{1} (\bibinfo{year}{2021}),
  \bibinfo{pages}{114--144}.
\newblock
\showDOI{%
\url{https://doi.org/10.1080/10556788.2020.1808977}}


\bibitem[\protect\citeauthoryear{Hansen, Finck, Ros, and Auger}{Hansen
  et~al\mbox{.}}{2009}]%
        {hansen:inria-00362633}
\bibfield{author}{\bibinfo{person}{Nikolaus Hansen}, \bibinfo{person}{Steffen
  Finck}, \bibinfo{person}{Raymond Ros}, {and} \bibinfo{person}{Anne Auger}.}
  \bibinfo{year}{2009}\natexlab{}.
\newblock \bibinfo{booktitle}{{\em {Real-Parameter Black-Box Optimization
  Benchmarking 2009: Noiseless Functions Definitions}}}.
\newblock \bibinfo{type}{Research Report} RR-6829.
  \bibinfo{institution}{{INRIA}}.
\newblock
\showURL{%
\url{https://hal.inria.fr/inria-00362633}}


\bibitem[\protect\citeauthoryear{Lacroix and McCall}{Lacroix and
  McCall}{2019}]%
        {limit}
\bibfield{author}{\bibinfo{person}{Benjamin Lacroix} {and}
  \bibinfo{person}{John McCall}.} \bibinfo{year}{2019}\natexlab{}.
\newblock \showarticletitle{Limitations of benchmark sets and landscape
  features for algorithm selection and performance prediction}. In
  \bibinfo{booktitle}{{\em Proceedings of the Genetic and Evolutionary
  Computation Conference Companion}}. \bibinfo{pages}{261--262}.
\newblock


\bibitem[\protect\citeauthoryear{Mu{\~n}oz and Smith-Miles}{Mu{\~n}oz and
  Smith-Miles}{2020}]%
        {munoz2020generating}
\bibfield{author}{\bibinfo{person}{Mario~A Mu{\~n}oz} {and}
  \bibinfo{person}{Kate Smith-Miles}.} \bibinfo{year}{2020}\natexlab{}.
\newblock \showarticletitle{Generating new space-filling test instances for
  continuous black-box optimization}.
\newblock \bibinfo{journal}{{\em Evolutionary computation\/}}
  \bibinfo{volume}{28}, \bibinfo{number}{3} (\bibinfo{year}{2020}),
  \bibinfo{pages}{379--404}.
\newblock


\bibitem[\protect\citeauthoryear{{Tian}, {Peng}, {Zhang}, {Rodemann}, {Tan},
  and {Jin}}{{Tian} et~al\mbox{.}}{2020}]%
        {tian_recommender_2020}
\bibfield{author}{\bibinfo{person}{Y. {Tian}}, \bibinfo{person}{S. {Peng}},
  \bibinfo{person}{X. {Zhang}}, \bibinfo{person}{T. {Rodemann}},
  \bibinfo{person}{K.~C. {Tan}}, {and} \bibinfo{person}{Y. {Jin}}.}
  \bibinfo{year}{2020}\natexlab{}.
\newblock \showarticletitle{A Recommender System for Metaheuristic Algorithms
  for Continuous Optimization Based on Deep Recurrent Neural Networks}.
\newblock \bibinfo{journal}{{\em IEEE Transactions on Artificial
  Intelligence\/}} \bibinfo{volume}{1}, \bibinfo{number}{1}
  (\bibinfo{year}{2020}), \bibinfo{pages}{5--18}.
\newblock
\showDOI{%
\url{https://doi.org/10.1109/TAI.2020.3022339}}


\bibitem[\protect\citeauthoryear{Škvorc, Eftimov, and Korošec}{Škvorc
  et~al\mbox{.}}{2020}]%
        {SKVORC2020106138}
\bibfield{author}{\bibinfo{person}{Urban Škvorc}, \bibinfo{person}{Tome
  Eftimov}, {and} \bibinfo{person}{Peter Korošec}.}
  \bibinfo{year}{2020}\natexlab{}.
\newblock \showarticletitle{Understanding the problem space in single-objective
  numerical optimization using exploratory landscape~analysis}.
\newblock \bibinfo{journal}{{\em Applied Soft Computing\/}}
  \bibinfo{volume}{90} (\bibinfo{year}{2020}), \bibinfo{pages}{106138}.
\newblock
\showISSN{1568-4946}
\showDOI{%
\url{https://doi.org/10.1016/j.asoc.2020.106138}}


\end{thebibliography}

\end{document}